\def\year{2020}\relax
\documentclass[letterpaper]{article} 
\usepackage{aaai20}  
\usepackage{times}  
\usepackage{helvet} 
\usepackage{courier}  
\usepackage[hyphens]{url}  
\usepackage{graphicx} 
\usepackage{graphicx}  
\usepackage{array,multirow}
\usepackage{enumitem}
\usepackage{amsmath}
\usepackage{amsfonts}
\usepackage{xcolor} 
\usepackage{booktabs}
\graphicspath{ {figs/} }

\newcommand{\newcite}[1]{\citeauthor{#1} \shortcite{#1}}

\title{Dual Reinforcement-Based Specification Generation for Image De-Rendering}
\author{
Ramakanth Pasunuru$^{\clubsuit}$ \hspace{4mm} David Rosenberg$^{\dagger}$ \hspace{4mm} Gideon Mann$^{\dagger}$ \hspace{4mm} Mohit Bansal$^{\clubsuit}$ \\
UNC Chapel Hill$^{\clubsuit}$ \hspace{4mm} Bloomberg LP$^{\dagger}$\\  
\texttt{\{ram,mbansal\}@cs.unc.edu} \\ \texttt{\{drosenberg44,gmann16\}@bloomberg.net}\\
}

\begin{document}

\maketitle

\begin{abstract}
Advances in deep learning have led to promising progress in inferring graphics programs by de-rendering computer-generated images. However, current methods do not explore which decoding methods lead to better inductive bias for inferring graphics programs.
In our work, we first explore the effectiveness of LSTM-RNN versus Transformer networks as decoders for order-independent graphics programs. Since these are sequence models, we must choose an ordering of the objects in the graphics programs for likelihood training.  We found that the LSTM performance was highly sensitive to the sequence ordering (random order vs. pattern-based order), while Transformer performance was roughly independent of the sequence ordering. Further, we present a policy gradient based reinforcement learning approach for better inductive bias in the decoder via multiple diverse rewards based both on the graphics program specification and the rendered image. We also explore the combination of these complementary rewards. We achieve state-of-the-art results on two graphics program generation datasets.
\end{abstract}

\section{Introduction}
\label{sec:intro}
The large majority of computer vision work deals in the domain of natural images
or video. However, there is tremendous potential for applying computer vision
techniques to computer-generated images, such as plots, charts, schematics, complicated
math formulas, and even a page of printed text. For these domains, there is often a
domain-specific language for precisely specifying the image, such as matplotlib
code for a chart, PicTeX for a schematic\footnote{\url{https://ctan.org/pkg/pictex?lang=en}},
and LaTeX for math formulas and text. ``De-rendering" a computer-generated image back 
to the original (or a different) domain-specific language specification can be a useful first step
in many tasks, such as changing the visual appearance of an
image~\cite{huang2016shape,wu2017neural}
or extracting information contained in an image~\cite{cliche2017scatteract,mishchenko2011chart}.

\begin{figure}
\centering
\includegraphics[width=0.9\linewidth]{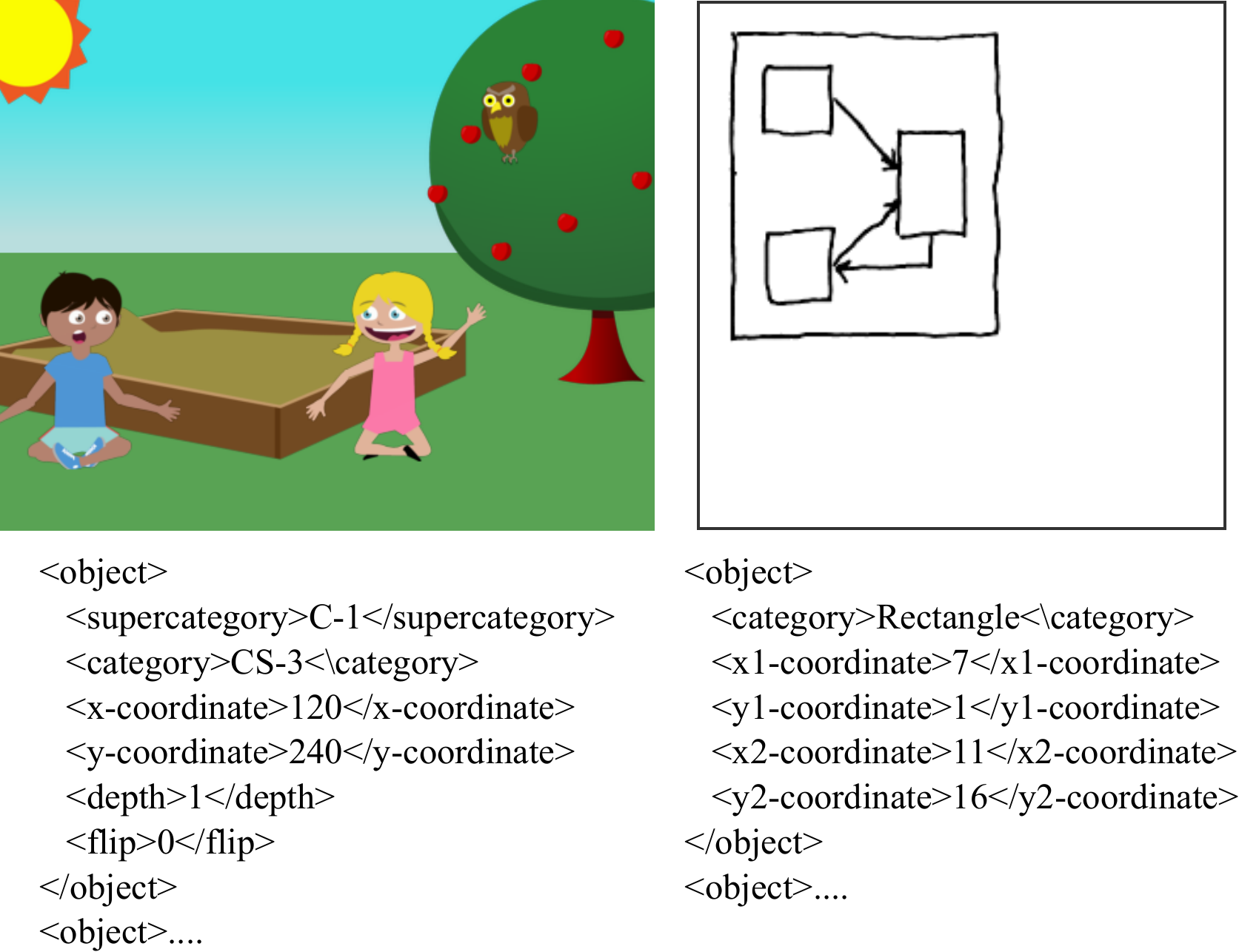}
\caption{Example images from the abstract scene dataset (left) and the Noisy Shapes dataset (right), along with portions of their specifications.
}
\label{fig:dataset_example}
\end{figure}

The de-rendering problem is part of a larger class of ``image-to-text"
problems, in which an input image is mapped to some sequence of output tokens.  
The neural encoder-decoder approach has proved to be very successful for this class of problems, including image captioning~\cite{karpathy2015deep,xu2015show}, handwriting recognition~\cite{DBLP:journals/corr/BlucheLM16}, as well as the
de-rendering problem for math formulas~\cite{deng2017image} and graphics images~\cite{ellis2017learning,wu2017neural}. In this paper, we improve these encoder-decoder
models for the specific case of graphical images, via methods based on Transformer models with both cross-entropy training and reinforcement learning with up to two ``dual modality'' reward functions.
De-rendering graphical images is a problem that differs in several
interesting ways from image captioning and OCR problems.  Two examples of the de-rendering problem we consider are shown in Fig.~\ref{fig:dataset_example}.  Each image is an input, and a portion of the desired output is displayed below each image.  In de-rendering, every object in the image must be described in the specification, and typically many output tokens are required to describe each object. Thus outputs from de-rendering are typically much longer than those in image captioning datasets~\cite{chen2015microsoft}, since caption labels (e.g., in COCO \cite{DBLP:journals/corr/LinMBHPRDZ14}) tend to focus on simple descriptions involving only the most salient objects in the image.
OCR and de-rendering are similar in that they encode information about all
elements of the image,
but the order of
the output sequence in OCR is completely determined by the image, while in
de-rendering, the output sequences represent sets\footnote{We say sets, rather than sequences, because in our datasets object ordering does not affect the rendering.}, and as
such the final rendering is invariant to a large degree of reordering in the
output sequence (e.g., by shuffling the sub-sequences of tokens that correspond to separate objects). 

We start our investigation with a basic image captioning model (similar to~\newcite{wu2017neural}) and extend it with an attention mechanism. We then swap out the LSTM-RNN decoder with a Transformer network~\cite{vaswani2017attention}. 
Our original motivation for this replacement is that we think that output generation requires long-term dependencies to avoid representing the same object multiple times.  As mentioned above, de-rendered output sequences can be quite long, and we thought the multi-head attention mechanism of the Transformer would handle the long-range dependencies better than the LSTM-RNN.  Unexpectedly, we found another advantage of Transformers over LSTM-RNNs for handling output sequence that can be reordered in many ways and still be correct.  We expand on this in Sec.~\ref{subsec:lstmvstransformer}.
To our knowledge, we are the first ones to use Transformer networks for de-rendering graphical images, and we find this change is a significant source of our performance improvement.

Another challenge with graphics de-rendering is that changing one or a few tokens in the specification can cause a significant change in many pixel values (e.g., by changing the location or color of a large object). Conversely, one can have two images that are very close visually, yet have completely different specifications. To this end, we explore the error minimization in the image as well as the specification space via a dual-modality, two-way reward reinforcement learning approach~\cite{williams1992simple,zaremba2015reinforcement}. We train with non-differentiable reward functions that reflect performance measures of interest in both the image space and the specification space (the ``dual modes'').
We further explore training a single model using rewards from both modalities, with the hopes that we get complementary feedback from each.

We empirically evaluate our methods on two image de-rendering datasets: Noisy Shapes dataset~\cite{ellis2017learning} and Abstract Scene dataset~\cite{zitnick2013bringing,wu2017neural}.  Our Transformer models trained with cross-entropy loss achieve very significant improvement over previous work on these datasets. We show even more improvement when we train the Transformer models using policy gradients-based methods, both via single-modality rewards and further improvements via dual-modality joint rewards. Finally, in our analysis we find evidence that the performance of Transformers is relatively insensitive to the ordering of objects in the output sequence,
while the performance of LSTM-RNN's can decay substantially for a poorly chosen object ordering.  This suggests that the advantage of Transformers over LSTM-RNNs may be particularly strong in tasks where we are using an output sequence to represent an unordered set of objects.


\section{Related Work}
\label{sec:related_work}
De-rendering a computer-generated image to a domain-specific language provides an abstraction that is easy to change, store, compare and match to other images. As a consequence, there has been recent interest and work in this area. 
\newcite{huang2016shape} used CNNs to translate a hand-drawn sketch of an object (e.g., jewellery) to a fixed set of parameters for a procedural model. In a similar vein,~\newcite{nishida2016interactive} proposed a simple procedural grammar as a building block to turn sketches into realistic 3D models.
\newcite{ellis2017learning} proposed an automatic visual program induction model to infer programs from hand-drawn images, where the images are encoded via CNNs and a multi-layer perceptron predicts a distribution over drawing commands. 

\newcite{ha2017neural} presented a recurrent neural network based sketch-rnn to construct conditional and unconditional sketch generation of common objects, constrained by a very simple set of primitives. Their model describes images as pen movements either in a drawing mode or in a non-drawing mode. Unlike our approach, this program is highly sequence dependent and non-compositional. While there are different solutions to the problem by re-ordering, one cannot arbitrarily shuffle the sequence of pen movements.~\newcite{liu2018learning} infer scene programs by exploiting hierarchical object-based scene representations.~\newcite{sun2018neural} proposed a neural program synthesizer that generates underlying programs for behaviorally diverse demonstration videos.
In this work, we use Transformer networks~\cite{vaswani2017attention} for decoding the specification from the given input image. Transformers have been used in other generation tasks such as image and video captioning~\cite{sharma2018conceptual,zhou2018end}, however, we are the first ones to use Transformer networks for the image de-rendering problem.~\newcite{vinyals2015order} shows that an LSTM trained with shuffled targets (unordered) using cross-entropy has a substantial drop in performance compared to natural orderings. Our result supports their findings and moreover we find that Transformers, by contrast, are relatively insensitive to the ordering of the objects.

Recently, policy gradient-based reinforcement learning (RL) methods have been widely used for sequence generation tasks: machine translation~\cite{ranzato2015sequence}, image captioning~\cite{ranzato2015sequence,rennie2016self}, and textual summarization~\cite{paulus2017deep,pasunuru2018multi}.~\newcite{halsearch} proposed to improve sequence generation by allowing a model to use its own prediction at training time, extending their work in structured prediction. In the context of program synthesis,~\newcite{bunel2018leveraging} used RL for generating semantically correct programs. 
In the context of image de-rendering,~\newcite{wu2017neural} proposed a neural scene de-rendering model (NSD) with a neural encoder and a graphics engine as a decoder. The encoder has an object proposal generator that produces segment proposals, and then it tries to interpret objects and their properties from these segments. They use RL to better sample the proposals and use the rendered image reconstruction error as reward.

Recently,~\newcite{ganin2018synthesizing} introduced an adversarially trained agent that is trained via a reinforcement learning setup without any supervision to generate a program that is executed by a graphics engine to interpret and sample images. In contrast, our work presents two complementary rewards (one in image space and another in specification space) in a reinforcement learning setup for the image de-rendering problem.

\section{Models}
\label{sec:models}

\paragraph{Task.}
For each task we consider, there is a simple graphics specification language that can be used to specify a particular image. While differing in details, the overall scheme of the specifications are the same for each. A specification consists of a set of ``objects'', and each object is specified by a set of properties. Examples of an object specification for each of our tasks can be seen in Fig.~\ref{fig:dataset_example}.  
Given an image rendered from a specification, our task is to ``de-render'' this image back to the original specification. We can evaluate a predicted specification by looking for exact matches between the objects in the predicted specification and the objects in the original specification.  We can summarize the object matches with standard  measures, such as precision, recall, F1, and intersection-over-union.  While these measures describe performance on a single image, we can average these measures across a collection of images, to get a performance measure of a method overall. We provide more details in Sec.~\ref{subsec:evaluation-metrics}.
Another approach to evaluation is to generate the image corresponding to a predicted specification, and see how well it matches the original image, using some reasonable metric on the space of images.

\noindent\textbf{Reduction to sequence prediction.}
While each specifications is represented by a set of objects with specific properties, our models require sequences of tokens.  We convert the set of objects to a sequence of tokens via some ordering of the objects. We investigate various approaches to ordering (Sec.~\ref{subsec:lstmvstransformer}), and find that ordering by object type works best.
Once the model predicts a sequence of tokens, we can parse it back into original structure to compute performance measures and our reward functions for reinforcement learning.

\subsection{Image-to-LSTM Sequence Model}
Our baseline model is similar to an image captioning model with
an attention mechanism~\cite{xu2015show}. We use the ResNet-18 architecture~\cite{he2016deep}
for encoding the input image, and we use an LSTM-RNN for predicting
the corresponding specification as a sequence of tokens.

We will denote the convolutional features from the ResNet-18 as $\{f_{i}\}_{i=1}^{m}$,
where $f_{i}\in\mathbb{R}^{d}$. For any decoder output token $o$, let $E_{o}\in\mathbb{R}^{d'}$
denote its embedding, which will be learned during training.
Let $s_{t}$ be the decoder state at step $t$, $o_{t}$ be the output
token at step $t$, and $c_{t}$ be the image context vector at step
$t$, which will be defined below. Then at step $t$, the
decoder state $s_{t}$ is given by
\begin{equation}
s_{t}=\mathbb{F}(c_{t},s_{t-1},E_{o_{t-1}}),
\end{equation}
where $\mathbb{F}$ is a trainable non-linear function. The context
vector $c_{t}$ is a convex combination of the image features: $c_{t}=\sum_{i=1}^{m}\alpha_{t,i}f_{i}$, where $\alpha_{t,i}$ are ``attention weights'' defined as
\begin{align}
\alpha_{t,i} &= \frac{\exp(e_{t,i})}{\sum_{k=1}^{m}\exp(e_{t,k})}\\
e_{t,i} &= v^{T}\tanh(Wf_{i}+Us_{t-1}+b),
\end{align}
where $v$, $W$, $U$, and $b$ are the trainable parameters.

\begin{figure}[t]
\centering
\includegraphics[width=0.8\linewidth]{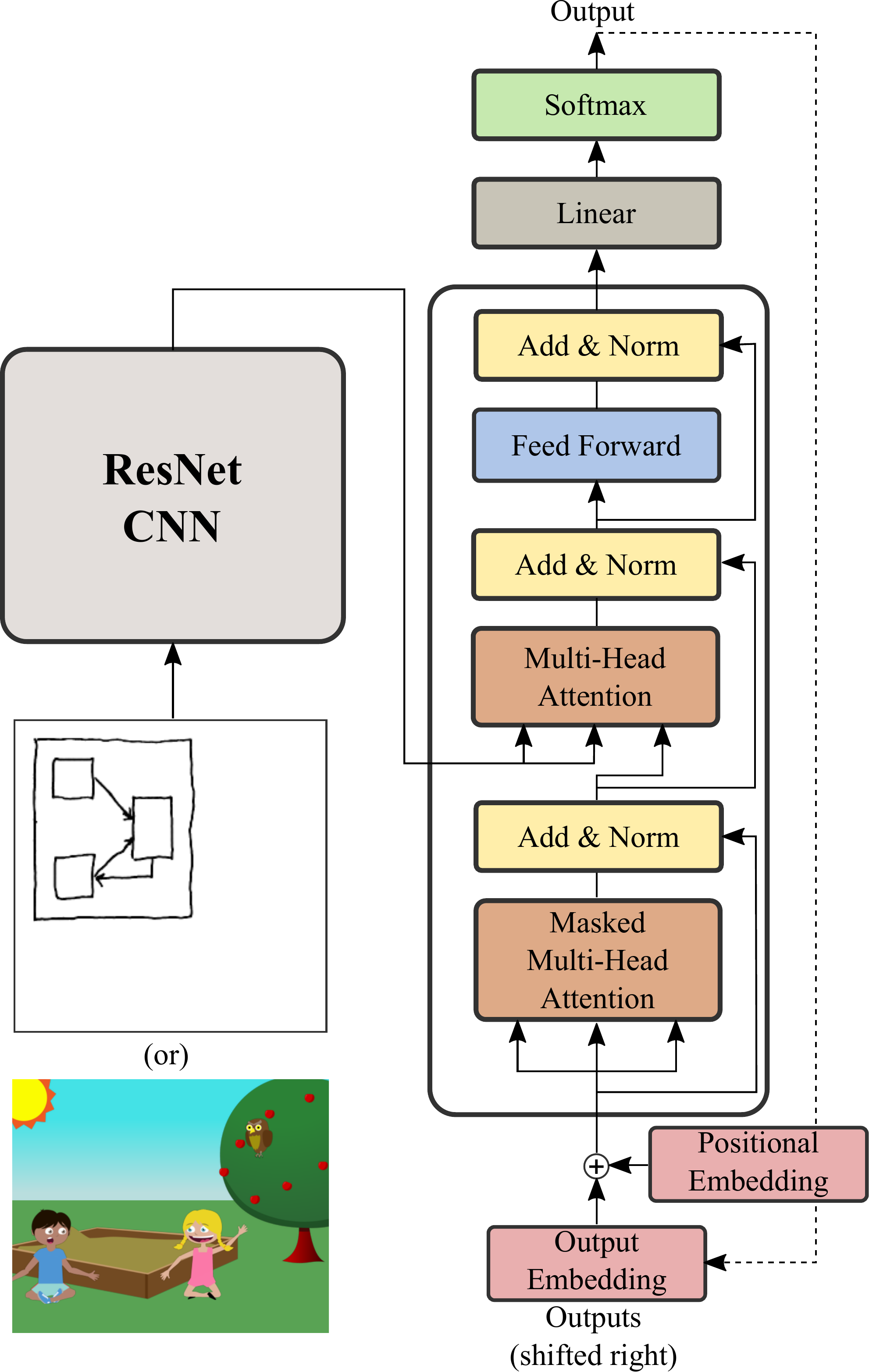}
\caption{Our Image-Transformer model.
}
\label{fig:main_model}
\end{figure}

\subsection{Image-to-Transformer Sequence Model} 
Recently, there is an increasing amount of interest in Transformer networks~\cite{vaswani2017attention}, which are said to train faster and to better capture long-term dependencies than LSTM-based RNN models. In our specification prediction problem, the length of the specification can be large, and we need long-term dependencies to avoid generating objects that have already been generated.  This suggests Transformer networks would be a better fit for our scenario.  In this work, we only use the decoder part of the Transformer network~\cite{vaswani2017attention}.  The Transformer encoder is for use on sequences, and we replace it with ResNet-18 CNN described above.  We give a high-level description of the Transformer decoder below, and refer to~\newcite{vaswani2017attention} for full details.

The decoder of the Transformer has a stack of $N$ identical layers containing self-attention modules, normalization modules, and feed-forward modules, along with positional encoding module for output embeddings (see Fig.~\ref{fig:main_model}). While the original model in~\newcite{vaswani2017attention} took $N{=}6$, through hyperparameter tuning we found $N{=}4$ to work better for our problem. Besides that, we used the hyperparameter settings as in~\newcite{vaswani2017attention}. The decoder has two attention modules: one for attending to the image convolution features and another self-attention module to attend to different previous positions in the decoder state. 

\paragraph{Attention in Transformer.}
As shown in Fig.~\ref{fig:main_model}, we have two attention mechanisms in the model: one attending to the CNN features, and another attending to different parts of the decoder state. They all have the same structure, which we describe below.

An attention mechanism in the Transformer can be viewed as a mapping from a query ($Q$) and a key-value ($K,V$) pair to an output. An attention weight is computed using the query and key and those weights are used with values to compute the output of the attention module. Empirically, it has been proven that instead of performing a single attention function, linearly projecting the queries, keys, and values with different learned projection layers and then performing the attention function in parallel and concatenating those outputs to get the final attention module output to work better. This attention mechanism is called multi-head attention mechanism (MH), which is defined as follows:
\begin{equation}
    \textrm{MH}(Q,K,V) = \textrm{Concat}(\textrm{head}_1,..,\textrm{head}_h)W^O
\end{equation}
\begin{equation}
  \textrm{head}_i = \textrm{Attention}(QW_i^Q, KW_i^K, VW_i^V)
\end{equation}
\begin{equation}
    \textrm{Attention}(Q,K,V) = \textrm{softmax}\left(\frac{QK^T}{d_k}\right) V 
\end{equation}
where, $d_k$ is the dimension of the queries and keys, $W_i^Q$, $W_i^K$, and $W_i^V$ are the parameters of the projection matrices.
\paragraph{Position-wise Feed-Forward Networks.}
In addition to the attention sub-layers, each of the layers in the Transformer decoder contains a fully connected feed-forward network that is applied to each position of the decoder separately and identically.  This network is defined as
\begin{equation}
    \textrm{FFN}(x) = \max(0, xW_1 + b_1)W_2 + b_2, 
\end{equation}
where $W_1$, $W_2$, $b_1$, and $b_2$ are the linear projection parameters which are same across different positions but are different from layer to layer. 

\paragraph{Positional Encoding.}
In the model described thus far, the model is symmetric with respect to sequence position.  For example, at the bottom right of Fig.~\ref{fig:main_model}, the model has no structural way to determine which output embeddings come from which part of the output sequence.  To remedy this issue, we concatenate a ``positional encoding'' ($\textrm{PE}$) to the embedding representation of the tokens. We use the sine and cosine functions for positional encoding:
\begin{equation}
\begin{split}
    \textrm{PE}(\textrm{pos}, 2i) = \textrm{sin}(\textrm{pos}/10000^{2i/d_{\textrm{model}}}) \\
    \textrm{PE}(\textrm{pos}, 2i+1) = \textrm{cos}(\text{pos}/10000^{2i/d_{\textrm{model}}}) 
\end{split}
\end{equation}
where $\textrm{pos}$ is the position, $i$ is the dimension, and $d_{\textrm{model}}$ is the dimension of the embedding vector representation.

\begin{figure*}
\centering
\includegraphics[width=0.86\linewidth]{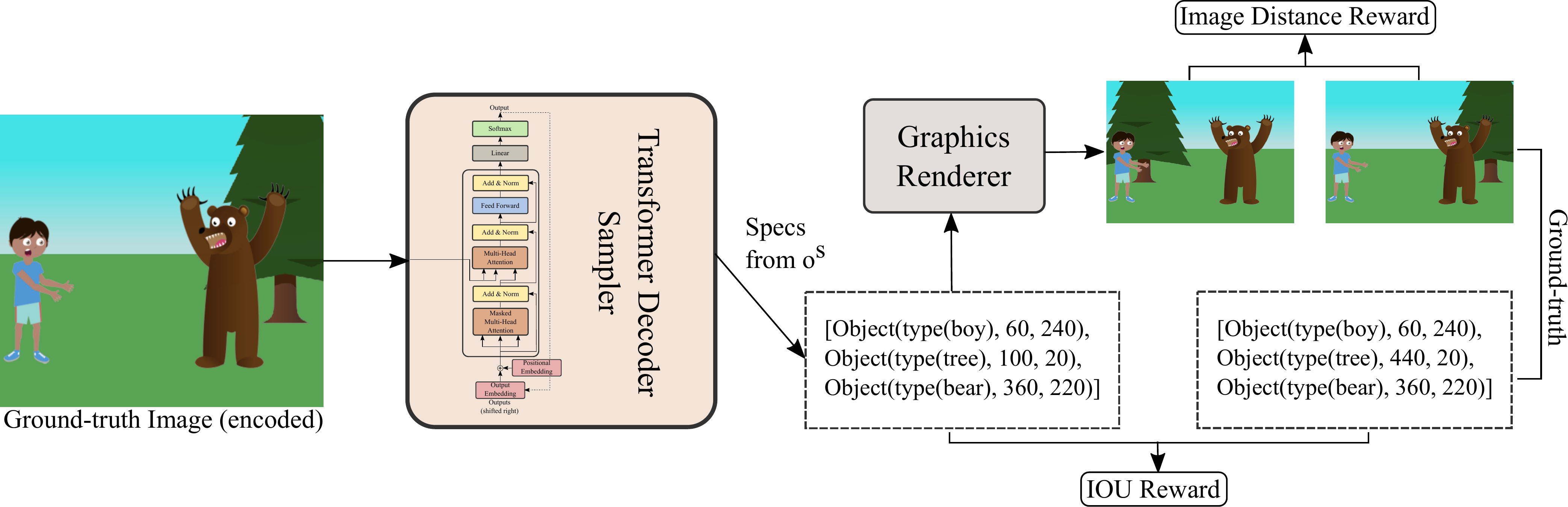}
\caption{Example showing the samples from our model based on abstract scene dataset and the corresponding rewards in specification and image space. For simplicity, all object properties are not shown in specification space. 
}
\label{fig:rl_model}
\end{figure*}

\section{Dual-Modality Two-Way Reinforcement Learning}
Traditionally, sequence generation models are trained using a cross-entropy loss. More recently, a policy gradient-based reinforcement learning approach has been explored for sequence generation tasks~\cite{ranzato2015sequence,rennie2016self}, which has two advantages over the cross-entropy loss optimization approach: (1) avoiding the \emph{exposure bias} issue, which is about the imbalance in the output distributions created by different train and test time decoding approaches in cross-entropy loss optimization~\cite{bengio2015scheduled,ranzato2015sequence}; (2) allows direct optimization of the evaluation metric of interest, even if it is not differentiable. To this end, we use a policy gradient-based approach via  rewards in both the specification space and the image space. Also, we explore joint rewards based on these two spaces for better capturing feedback that is complementary between these two modalities.

For this reward optimization, we use the REINFORCE algorithm~\cite{williams1992simple,zaremba2015reinforcement} to learn a policy $p_{\theta}$ that produces a distribution over sequences $o^s$ for any given input. We try to find a policy $p_{\theta}$ such that the expected reward for a label sequence $o^s$ drawn according to the predicted distribution has maximum expected reward. Equivalently, we minimize the following loss function, in average across all training inputs:
\begin{equation}
    L_{\textrm{RL}} = -\mathbb{E}_{o^s \sim p_\theta} [r(o^s)],
\end{equation}
where $o^s$ is the sequence of sampled tokens with $o^s_t$ sampled at time step $t$ of the decoder.  We can approximate the gradient of this loss function with respect to the parameter $\theta$ using a single sample $o^s$ drawn from $p_\theta$ as:
\begin{equation}
\nabla_\theta L_{\textrm{RL}} = -(r(o^s)-b_e) \nabla_\theta \log p_\theta(o^s),
\end{equation}
where the leading factor is included for variance reduction using a baseline
estimator~\cite{zaremba2015reinforcement}. There are several ways to calculate the baseline estimator; we employ the effective SCST approach~\cite{rennie2016self}.

\subsection{Rewards}
In this work we consider three different reward functions.  Two of the rewards are based in ``specification space'', which make a direct comparison between the predicted specification and the ground truth specification, and one of the rewards is based in ``image space'', which compares the image rendered from the predicted specification with original input image.  We also investigate using these rewards in combination, with the hope that there is complementary information in the feedback based on the two spaces.  

\subsubsection{Intersection-Over-Union Reward (IOU)}\label{iou-reward}
As mentioned in Sec.~\ref{sec:models}, after the specification is predicted as a sequence of tokens, we can parse the sequence into a set of object specifications. The intersection-over-union (IOU) reward is based in specification space.  Roughly speaking, the IOU reward gives credit for predicting objects that exactly match objects in the ground truth specification, and penalizes both for predicting objects that do not match ground truth objects and for failing to predict objects that are part of the ground truth.  More formally, 
let $\{o_i\}_{i=1}^m$ and $\{o^*_j\}_{j=1}^n$ represent the objects in predicted and ground-truth specifications, respectively. Then the IOU reward is defined as:
\begin{equation}
    r_{iou} = \frac{\textrm{count}(\{o_i\}_{i=1}^m \cap \{o^*_j\}_{j=1}^n)}{\textrm{count}(\{o_i\}_{i=1}^m \cup \{o^*_j\}_{j=1}^n)}
\end{equation}
The object $o_i$ in the prediction specification is the same as object $o^*_j$ in the ground-truth specification if and only if all the properties of these objects match exactly. 

\subsubsection{Inference Reward}\label{inference-reward}
Our second reward, which we call the ``inference reward'', is also a reward in specification space.  The name is based on the ``inference error'', which is a performance measure introduced in~\newcite{wu2017neural} for the Abstract Scenes dataset.  While IOU is based on exact matches between predicted objects and ground-truth objects, the inference error and inference reward are based on the number of properties (within objects) that correctly match the corresponding properties in the ground-truth.  For those properties specifying location in pixel coordinates, we follow~\newcite{wu2017neural} and divide the space of each coordinate into 20 bins of equal size, and we consider it a match if the predicted and ground-truth locations are in the same bin. We define the inference error as the fraction of predicted properties that fail to match the corresponding ground-truth properties.  The inference reward is one minus the inference error.   

\subsubsection{Image Distance Reward}\label{image-distance-reward}
Our third and final reward, the ``image distance reward'', is in image space.  We define it generically first, as it takes slightly different forms in our two datasets.  If we let $I$ and $I^R$ represent vectorized versions of the input image and the image rendered from the predicted specification, respectively,  then we define the image distance as
\begin{equation}
    d_{img} = ||I \ominus \Psi(I^R)||^2_2,
\end{equation}
where $||\cdot||_2$ is the $\ell_2$-norm.

For Noisy Shapes dataset, we follow~\newcite{ellis2017learning} and take $\Psi$ to be a Gaussian blurring function, as the objects in the target image have noise (see Fig.~\ref{fig:dataset_example}).  We take $\ominus$ to be a simple subtraction operation. The image reward for this dataset is: 
\begin{equation}
r_{img} = \frac{c}{d_{img}}
\end{equation}
where $c$ is a tunable parameter. 

For the Abstract Scene dataset,  $\ominus$ is a logical operator that takes the value $0$ in every position where the pixel values ``match'', and $1$ in every other position.  The range of possible pixel values is 0-255 and, similarly to the discretization of position in the inference reward, we divide the pixel value range into 20 equisized buckets and consider pixel values to match if they are in the same bucket.  We take $\Psi$ to be the identity function.
The image reward for the Abstract Scenes dataset is then defined as: 
\begin{equation}
    r_{img} = 1 - \frac{d_{img}}{w \cdot h}
\end{equation}
where $w$ and $h$ are width and height of the image.

\subsubsection{Joint Dual-Modality Reward}
Since we expect the rewards based in specification space to be complementary to the reward based in image space, we want a way to combine rewards on the two spaces. One way to combine two rewards is to create a weighted combination of individual rewards to formulate the joint reward.  Another approach is to alternate the reward used during the learning process~\cite{pasunuru2018multi}. In this work, we follow the latter approach, as the former approach requires expensive tuning for scale and weight balancing. 
Let $r_1$ and $r_2$ be the two reward functions that we want to optimize. In our approach, we first take 
$a_1$ optimization steps to minimize the reinforcement learning loss $L_{\text{RL}_1}(r_1;\theta)$ (i.e. we use $a_1$ mini-batches). Then we take $a_2$ optimization steps to minimize the reinforcement learning loss $L_{\text{RL}_2}(r_2;\theta)$. We then repeat this cycle of steps until convergence.  All other optimization parameters, such as step size, remain the same for each set of steps. The values $a_1$ and $a_2$ are tuning parameters.\footnote{\newcite{pasunuru2018multi} set $a_1$ and $a_2$ to $1$, without tuning.} The two rewards $r_1$ and $r_2$ could be based on different aspects of the output, such as IOU and image distance reward.

\begin{table*}[t]
\begin{center}
\begin{tabular}{l|c|c|c|c|c|c|c}
\toprule
Model &  Precision & Recall & F1 & IOU & $\text{IOU}_{\text{1.0}}$ & $\text{IOU}_{\text{0.8}}$ & $\text{IOU}_{\text{0.6}}$ \\
\midrule
\multicolumn{8}{c}{\textsc{Cross-Entropy Loss}} \\
\midrule
Image2LSTM+atten.  & 98.7 &  98.5 & 98.6 & 97.6 & 90.7 & 95.3 & 98.8 \\
Image2Transformer & 99.1 & 99.1 & 99.1 & 98.5 & 94.1 & 97.3 & 99.1 \\
\midrule
\multicolumn{8}{c}{\textsc{Image2Transformer with Reinforce Loss}} \\
\midrule
IOU Reward & \textbf{99.4} & \textbf{99.3} & \textbf{99.3} & \textbf{98.8} & \textbf{95.0} & 98.0 & 99.4 \\
Image-distance Reward  & \textbf{99.4} & 99.2 & \textbf{99.3} & \textbf{98.8} & 94.5 & 98.0 & \textbf{99.5} \\
Image-distance + IOU Reward  & \textbf{99.4} & \textbf{99.3} & \textbf{99.3} & \textbf{98.8} & \textbf{95.0} & \textbf{98.1} & 99.4 \\
\bottomrule
\end{tabular}
\end{center}
\caption{Performance of various models on Noisy Shapes dataset.
}
\label{table:main-results}
\end{table*}

\section{Experimental Setup}

\subsection{Dataset}

\paragraph{Noisy Shapes Dataset.}
\newcite{ellis2017learning} provides a synthetic dataset of images containing multiple simple objects (lines, circles, and rectangles), each with various properties that can be specified. The images are specified using a small subset of $\text{\LaTeX}$ drawing commands.  Additional noise is introduced into the rendered images by rescaling image intensity, translating the image by a few pixels, rendering the $\text{\LaTeX}$ using the $\operatorname{pencildraw}$ style, and randomly perturbing the position and sizes of these $\text{\LaTeX}$ drawing commands. The dataset was created by randomly sampling image specifications with between 1 and 12 objects, excluding any specifications that lead to images with overlapping objects. The size of each image is 256x256.
The dataset contains 100,000 images paired with specifications, from which we use 1000 for testing and the rest for training. 

\paragraph{Abstract Scene Dataset.}
The Abstract Scene dataset~\cite{zitnick2013bringing} contains 10,020 images, each of which has 3-18 objects. There are over 100 types of objects, each of which is specified by two integers, one indicating a broad category (e.g. sky object, animal, boy, girl) and another indicating a subcategory (e.g. girl pose, animal type, etc.). Each object can be drawn at one of 3 scales, with or without a horizontal flip, and at any pixel location in the 500x400 image.  These properties are specified by 4 additional integers.  Thus each object is specified by 6 integers.  There are often heavy occlusions among these objects when rendered in an image (see input image in Fig.~\ref{fig:main_model}).  However, the objects are rendered in a deterministic order based on the object types and other properties, and thus the image is independent of the order of the objects in the specification. Similar to~\newcite{wu2017neural}, we randomly sample 90\% of the images for training and rest for testing. 

\subsection{Evaluation Metrics}
\label{subsec:evaluation-metrics}

\paragraph{Noisy Shapes Dataset.}
As described in the Task description of Sec.~\ref{sec:models}, we can summarize performance on a single image with precision, recall, F1, and IOU (intersection over union) at the object level. Following previous work~\cite{ellis2017learning}, we summarize the performance of a method by averaging these metrics across all test examples (i.e. a macro average).
Further, we also report $\text{IOU}_k$, which is defined as the percent of test examples for which the IOU score is greater than or equal to $k$.

\paragraph{Abstract Scene Dataset.}
For the abstract scene dataset, following previous work~\cite{wu2017neural}, we report specification inference error and image reconstruction error based on a micro average across all test examples. As described in Sec.~\ref{inference-reward} and Sec.~\ref{image-distance-reward}, inference error is based on the percentage of incorrectly inferred values (i.e., how many properties of objects do not match with the ground-truth) for the specification, and image reconstruction error is based on percentage of incorrect pixel prediction. During these evaluations, all the continuous variables (pixel values, and x and y coordinates) are quantized into 20 bins. Additionally, we report the macro average based IOU error as described for the noisy shapes dataset.

\subsection{Training Details}
In all of our models, we encode the image information via ResNet-18~\cite{he2016deep}, where we take the penultimate layer's features as outputs from this image encoder. For LSTM-RNN, we use a hidden state size of 128, input token embedding size of 128, and a batch size of 64. For Transformer networks, we use the same hidden and embedding size, and use 4 layers at each time step. We use the Adam optimizer~\cite{kingma2014adam} with the default learning rate of 0.001 for all the cross-entropy models, and a learning rate of 0.0001 for all the reinforcement learning based models. For the Noisy Shapes dataset, the maximum decoder length is fixed to 80, and we use a vocabulary size of 27, which are placeholders for object properties. For the Abstract Scene dataset, the maximum decoder length is fixed to 100, and we use a vocabulary size of 1078 which represents all the object properties. For the joint reward optimization, we use a mixing ratio of 1:1 for the Noisy Shapes dataset and 1:4 for the Abstract Scene dataset.

\section{Results}

\subsection{Results on the Noisy Shapes Dataset}

We first compare the performance of the LSTM-RNN model (Image2LSTM+atten) to the Transformer-based model, when both are trained with cross-entropy loss.  We see in Table~\ref{table:main-results} that the Transformer model dominates on all measures.  In particular, we highlight $\text{IOU}_{\text{1.0}}$, which measures the percent of examples on which the predicted specification exactly matches the ground-truth specification.  While the LSTM-RNN model achieves a $90.7\%$ $\text{IOU}_{\text{1.0}}$, the Transformer model achieves $94.1\%$, which is an impressive $36.5\%$ reduction in the number of errors.  We have similar performance improvements for the other metrics.
We now compare the Transformer model trained with reinforcement learning, using various reward functions, to training using cross-entropy loss.  Table~\ref{table:main-results} shows that, although all three reward variations have roughly the same performance, they all show significant improvement over cross-entropy training, on all measures.\footnote{The improvement of our Transformer models trained with reinforcement learning over the corresponding cross-entropy models is statistically significant with $p<0.01$, based on the bootstrap test~\cite{noreen1989computer,efron1994introduction}. } For example, the model trained with IOU reward achieved a $95.0\%$ $\text{IOU}_{\text{1.0}}$ measure, which is an impressive $15.3\%$ reduction in the number of errors compared to the same model trained with cross-entropy loss, and a $46.2\%$ reduction compared to the original LSTM-RNN model. Performance improvement in the other measures is at least as good.

\begin{table}[t]
\small
\begin{center}
\begin{tabular}{l|p{0.9cm}|p{0.9cm}|p{0.9cm}|p{0.9cm}}
\toprule
Model &  Infer. Error & Recons. Error & Avg. Error & IOU  \\
\midrule
\multicolumn{5}{c}{\textsc{Previous Work}} \\
\midrule
CNN+LSTM~\shortcite{wu2017neural}  & 45.31 &  41.38 & 43.84 & -  \\
NSD (full)~\shortcite{wu2017neural}  & 42.74 &  21.55 & 32.14 &  -  \\
\midrule
\multicolumn{5}{c}{\textsc{Cross-Entropy Loss}} \\
\midrule
Image2LSTM+atten.  & 17.27 & 15.70 & 16.48 & 32.06 \\
Image2Transformer & 8.78 & 10.92 & 9.85 & 58.54 \\
\midrule
\multicolumn{5}{c}{\textsc{Image2Transformer with Reinforce Loss}} \\
\midrule
IOU Reward & 7.91 & 10.50 & 9.20 & 61.29 \\
Inference Reward  & \textbf{7.81} & 10.75 & 9.28 & 59.35 \\
Recons. Reward & 8.34 & \textbf{9.99} & 9.16 & 62.44 \\
Inference + Recons. & 8.21 & 10.12 & 9.16 & 61.54 \\
IOU + Recons.  & 8.05 & 10.04 & \textbf{9.04} & \textbf{62.45} \\
\bottomrule
\end{tabular}
\caption{Models performance on the abstract scene dataset. Errors: lower is better; IOU: higher is better.
}
\label{table:abd-main-results}
\end{center}
\end{table}

\subsection{Results on Abstract Scene Dataset}
In  Table~\ref{table:abd-main-results}, we see the performance of various models on the Abstract Scene dataset, for the metrics described in Sec.~\ref{subsec:evaluation-metrics}. We first note that even our baseline LSTM-RNN model (Image2LSTM+atten) shows a very large error reduction compared to the results presented in \newcite{wu2017neural} (first 4 rows of the table). This highlights the importance of an attention mechanism in these tasks. For the models trained with cross-entropy, the Transformer model shows an additional remarkable improvement over the LSTM-RNN model, across all measures.  

For reinforcement learning with the Transformer model, we tried three different reward functions, corresponding to three of our performance metrics: inference error, reconstruction error, and IOU.  All the Transformer models trained with REINFORCE out-performed the model trained with cross-entropy loss for each of the error measures.\footnote{For the IOU and inference reward models, this improvement is statistically significant for all metrics except reconstruction error. For the reconstruction reward model, the improvement is significant for all but the inference error metric.  For the dual (IOU+Recons.) reward model, the difference is significant for all metrics ($p<0.01$ for each test).    
}
For inference error, the model trained with the inference reward did the best, as one might hope and expect.  Compared to the cross-entropy trained Transformer, the inference error measure was reduced by $11.0\%$.  For reconstruction error (image-based), the best performing model was the model trained with the reconstruction reward, which reduced the reconstruction error by $8.5\%$ compared to the cross-entropy trained version.  When evaluating performance using the average of the inference and reconstruction error, one of our joint-reward models performed best, though interestingly, not the one that uses the corresponding inference and reconstruction rewards.  The best performing model for this performance measure used IOU and reconstruction rewards, suggesting that IOU reward has more information that is complementary to the reconstruction error than does the inference reward. For IOU performance measure, the model trained with IOU reward did well, but when trained jointly with IOU and reconstruction reward, it performed the best. This suggests that using image-based feedback during training (recons. error) can be beneficial even when the ultimate goal (IOU) depends only on the specification output.

\section{Analysis}\label{}

\subsection{LSTM vs. Transformer Networks}
\label{subsec:lstmvstransformer}
As noted above, for the Abstract Scene and the Noisy Shapes datasets that we consider, the order of the objects in the specification does not affect the final image.  Nevertheless, for training both the LSTM-RNN and the Transformer models, one must choose an ordering.  We ran an experiment using the Noisy Shapes dataset, in which we tried ordering the objects by shape size, shape type, and by shape position in the rendered image.  We found that ordering by shape type worked best across our models, so that's what we used for our main results in Table~\ref{table:main-results}.  We also wanted to investigate how important it is to have the objects in some sensible order, compared to a random ordering.  Table~\ref{table:unordered-results} shows the results of our two models when trained with cross-entropy on specification sequences where the objects are put in random order.  We find that the LSTM-RNN model performance drops dramatically (e.g. $\text{IOU}_{\text{1.0}}$ drops from $90.7\%$ to $72.0\%$), while the drop with Transformer networks is quite small (e.g. $\text{IOU}_{\text{1.0}}$ drops from $94.1\%$ to $93.2\%$.\footnote{Note that the number of model parameters is approximately the same (11.8M for Transformer model and 11.5M for LSTM model). Further, Transformer models are 2.5x faster to train in comparison to the LSTM models. During inference, both models take approximately the same time.}  This is additional evidence for Transformers being the preferred model for tasks of this type.

\subsection{Performance vs. Data Size}

We conduct an experiment where we vary the percentage of Noisy Shapes data used during our models' training from $10\%$ to $100\%$ by steps of $20\%$. We observe that with less data ($10\%$-$40\%$), the RL-based model is approximately $2$ points better (on the $\text{IOU}_{\text{1.0}}$ metric) than its corresponding cross-entropy baseline. As we use more data (${>}60\%$), the gap decreases to $1$ point between RL and cross-entropy models. This suggests that RL, which has the advantage of exploration, is more powerful when the data is less.

\begin{table}
\begin{center}
\begin{tabular}{l|c|c|c}
\toprule
Model &  F1 & IOU & $\text{IOU}_\text{1.0}$  \\
\midrule
Image2LSTM+atten.  & 95.2 & 92.0 & 72.0 \\
Image2Transformer & 99.0 & 98.3 & 93.2 \\
\bottomrule
\end{tabular}
\end{center}
\caption{Performance of LSTM-RNN and Transformer networks on the Noisy Shapes dataset when specifications have randomly ordered objects.
}
\label{table:unordered-results}
\end{table}

\subsection{Output Examples}
\begin{figure}[t]
\centering
\includegraphics[width=0.9\linewidth]{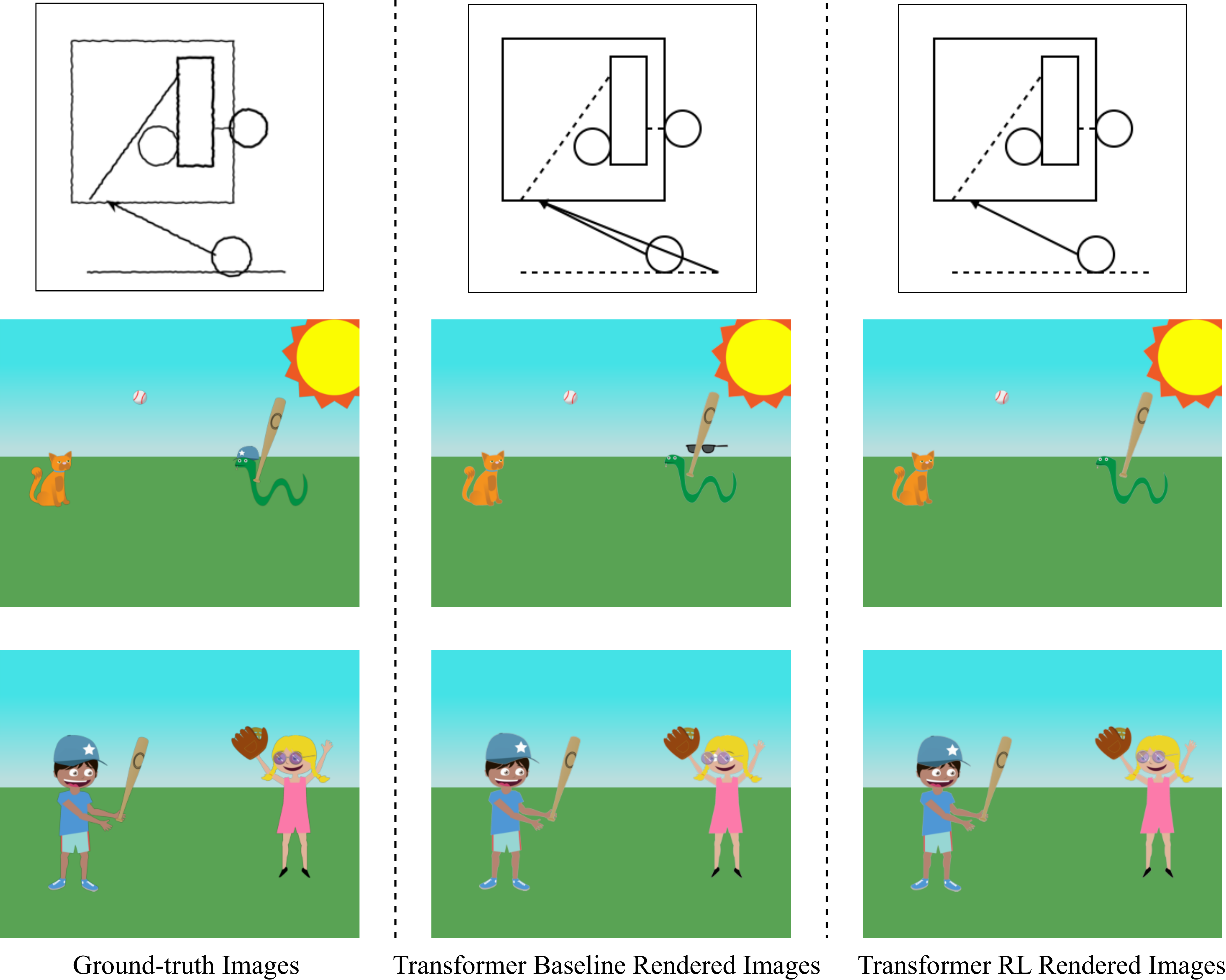}
\caption{Comparing the ground-truth images from noisy shapes and abstract scene datasets with the rendered images of predicted specifications.
}
\label{fig:output-examples}
\end{figure}

Fig.~\ref{fig:output-examples} presents the output rendered images of the predicted specifications from Image2Transformer cross-entropy model and the corresponding RL-based model with IOU+Image-distance as reward for noisy shapes dataset and IOU+Recons. as reward for abstract scene dataset. In the first example (top row in Fig.~\ref{fig:output-examples}), the cross-entropy model predicts an extra `line shape' which is not present in the ground-truth. Our RL model correctly predicts the exact same shapes present in the ground-truth. However, neither models getting the type of `line shape' correct in couple of instances. In the second example (second row in Fig.~\ref{fig:output-examples}), the cross-entropy model predicts an extra object (glasses), which is not present in the ground-truth image, and is also missing the cap on the snake. The RL model improves on the cross-entropy model by not having any extra objects, but it is also missing the cap.  In the third example, both the rendered images look very similar to the ground-truth, but the cross-entropy model predicts one of the objects (glasses) slightly off in position. Our RL model was able to accurately position the glasses (bottom row in Fig.~\ref{fig:output-examples}). The better performance of RL model may be due to the image space component of the error signal, which is more sensitive to position errors, while the cross-entropy loss gives the same penalty to all incorrect positions regardless of the error size.

\section{Conclusion}
We present various neural de-rendering models based on LSTMs with attention mechanism and Transformer networks. Further, we introduce complimentary dual rewards (one in specification space and another in image space) and optimize them via reinforcement learning, and achieve state-of-the-art results. Further, our results and analyses suggest that Transformers are a better choice than LSTMs for unordered sequence prediction tasks.

\section*{Acknowledgments}
We thank the reviewers for their helpful comments. This work was partially supported by NSF-CAREER Award 1846185, ARO-YIP Award W911NF-18-1-0336, and a Microsoft PhD Fellowship. The views contained in this article are those of the authors and not of the funding agency.

\bibliography{citations.bib}
\bibliographystyle{aaai}

\end{document}